\pdfoutput=1
\documentclass[sigconf]{acmart}
\AtBeginDocument{%
  }

\setcopyright{acmlicensed}
\copyrightyear{2018}
\acmYear{2018}
\acmDOI{XXXXXXX.XXXXXXX}
\acmConference[Conference acronym 'XX]{Make sure to enter the correct
  conference title from your rights confirmation email}{June 03--05,
  2018}{Woodstock, NY}
\acmISBN{978-1-4503-XXXX-X/2018/06}

\author{Yulin Wang}
\affiliation{%
  \institution{College of Computer and Information Science, \\ Southwest University}
  \city{Chongqing}
  \country{China}
}
\email{wyl72229992@email.swu.edu.cn}

\author{Yi He}
\affiliation{%
  \institution{Department of Data Science, \\ College of William and Mary}
  \city{Williamsburg}
  \state{VA}
  \country{USA}
}
\email{yihe@wm.edu}

\author{Dianlong You}
\affiliation{%
  \institution{School of Artificial Intelligence (School of Software), \\ Yanshan University}
  \city{Qinhuangdao}
  \state{Hebei}
  \country{China}
}
\email{youdl@ysu.edu.cn}

\author{Di Wu}
\authornote{Corresponding author.}
\affiliation{%
  \institution{College of Computer and Information Science, \\ Southwest University}
  \city{Chongqing}
  \country{China}
}
\email{wudi1986@swu.edu.cn}
\usepackage[linesnumbered,ruled,vlined]{algorithm2e}
\usepackage{multirow}
\usepackage{colortbl}  % 用于表格着色
\usepackage{xcolor}    % 颜色支持
\usepackage{graphicx}
\usepackage{subcaption}
\usepackage{amsmath} 

\usepackage{amsthm} % 如果您还没有定义 lemma 环境
\definecolor{mygray}{gray}{0.92} % 定义一种极淡的灰色，打印出来也很干净
\begin{document}

\title{DT-GOL: Dual-Track Geometric Online Learning in Nonstationary Environment with Label Delay}


\renewcommand{\shortauthors}{Trovato et al.}

\begin{abstract}
Online learning is crucial for handling complex data streams in big data applications. Recent research has begun to focus on dynamic scenarios, i.e., non-stationary environments. However, a crucial yet often overlooked aspect is label latency, where new data may not receive labels in time due to the slow and expensive labeling process, thus hindering rapid adaptation to dynamic environments.

To resolve this impasse, we propose Dual-Track Geometry Online Learning (DT-GOL), a novel framework that shifts from temporal compensation to spatial reasoning to bridge the supervised latency gap. By modeling the delay challenge as a semi-supervised task, we leverage real-time topological evolution of features as a reliable geometric surrogate for unobservable conceptual changes to achieve proactive supervised adaptation within the delay window. Unlike rigid self-training, we introduce a dynamic evidence calibration mechanism that distills geometric information into soft labels that perceive uncertainty, effectively mitigating the confirmation bias inherent in hard pseudo-labels. Furthermore, to resolve the stability-plasticity dilemma, we design a decoupled dual-track architecture in which a master learner serves as a stable anchor, updated strictly from delayed ground truth, while a transient branch leverages soft geometric knowledge for low-risk forward adaptation. Extensive experiments on real and synthetic datasets demonstrate that DT-GOL significantly outperforms existing state-of-the-art baseline methods, especially in scenarios with concept drift.
\end{abstract}

\begin{CCSXML}
<ccs2012>
 <concept>
  <concept_id>00000000.0000000.0000000</concept_id>
  <concept_desc>Online learning, Stream learning</concept_desc>
  <concept_significance>500</concept_significance>
 </concept>
</ccs2012>
\end{CCSXML}

\ccsdesc[500]{Online learning~Stream data}

\keywords{Do, Not, Use, This, Code, Put, the, Correct, Terms, for,
  Your, Paper}

\maketitle

\section{INTRODUCTION}

Online learning~\cite{bhatia2020online, he2023towards,10.1016/j.neucom.2021.04.112} has established itself as a fundamental paradigm for processing high-velocity data streams. In contrast to traditional batch learning methods that require retrospective access to complete datasets, online learning incrementally refines model parameters in real-time as new instances arrive. This capability significantly reduces computational and storage overhead while delivering exceptional adaptability for intricate real-world applications, such as recommendation systems~\cite{10.1109/TKDE.2025.3544510,koren2009matrix,luo2021fast,wu20211,zhang2023lightfr}, privacy-preserving federated services~\cite{guendouzi2023systematic,li2020review,banabilah2022federated,lin2020meta,ivannikova2019federated,chai2020secure,lin2020fedrec,rahimi2023evofed,zhu2019deep,gao2025federated,wu2026federated,yu2026federated}, structured data interfaces~\cite{wu2026schemarag}, high-frequency trading~\cite{10.5555/3042573.3042648}, and outlier detection~\cite{11117179}.

To enhance practical deployment capabilities, a growing body of research on online learning in nonstationary environments has emerged~\cite{10.1145/2523813,wang2024non}. This environment poses critical challenges, such as concept drift~\cite{8496795,10.1145/3472752}, feature heterogeneity~\cite{dekel2008learning}, feature incompleteness~\cite{zhao2020missing}, and high-dimensional sparse representation~\cite{wu2023mmlf,wu2020data,9885025,10179251,wu2026multimetric,wu2026non}. Related latent factor and tensor representation studies further highlight the importance of robust modeling under sparse, incomplete, and noisy data~\cite{9783168,wu2023robust,ma2025review,yu2025multi}. Despite existing methodologies having made notable strides in mitigating these issues individually or in pairs~\cite{zhuo2025online,10.1109/TKDE.2024.3374357,10.1109/TKDE.2023.3250472}, they predominantly rely on the assumption that the true label of each data point becomes available immediately after prediction, enabling continuous model refinement through instantaneous feedback. In practical applications, label acquisition significantly lags behind feature arrival, leading to unavoidable verification latency~\cite{he2019practical,dunn2021wearable}, as the labeling process is often slow, expensive, and asynchronous. Combined with complex streaming data, we refer to this scenario as \textbf{Nonstationary Environments with Label Delay}.

\begin{figure}[tt]
  \centering
  \includegraphics[width=\linewidth]{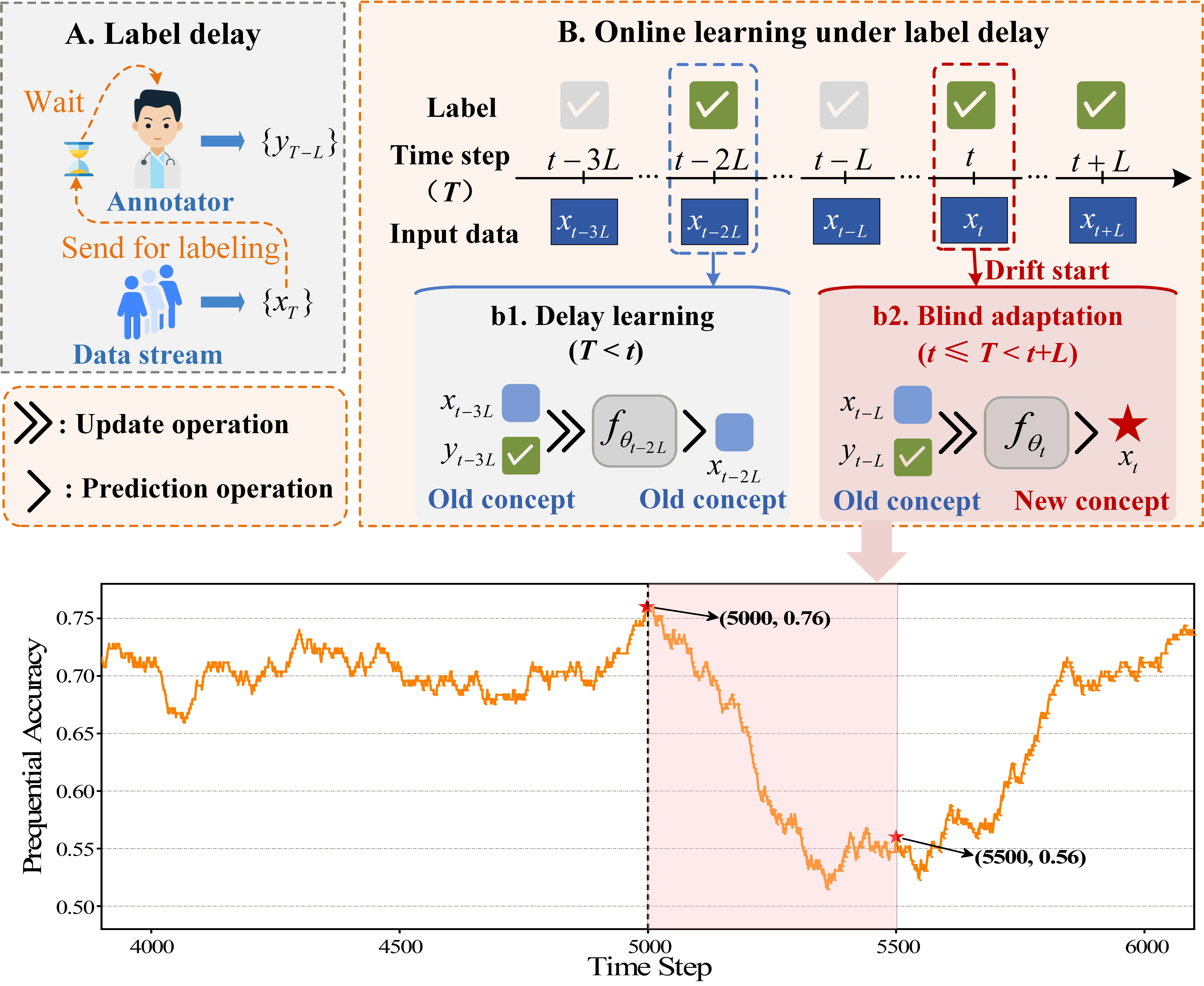}
  \caption{Illustration of label delay. (A) illustrates the basic concept of validation delay. At each time step $T$, the annotator reveals a label $y_{T-L}$ that arrives $L$ steps late, and then the data stream reveals an unlabeled data $x_T$ for evaluation, whose true label arrives $L$ steps late. (B) illustrates that encounter two challenges under label delay: delay learning and blind adaptation. The below shows performance trends for $L$=500, where the concept drift starts at 5000. The pink shaded area represents the intersection of the new concept and label delay interval, $i.e.$, blind adaptation zone.}
  \label{introduction}
\end{figure}

A quintessential example is postoperative rehabilitation monitoring. As illustrated in Figure~\ref{introduction}(A), physicians collect patient data daily, but diagnostic labels such as "recovery" or "complications" require weeks of evaluation to be confirmed. Such creates a severe temporal misalignment: critical treatment decisions rely on real-time patient physiological state ($x_T$), whereas the model guiding these decisions can only be updated using historical data pairs $(x_{T-L}, y_{T-L})$ from weeks earlier. An inevitable performance gap of $L$ steps consequently arises relative to immediate feedback scenarios. More dangerously, if concept drift intersects within this delay window, as illustrated in Figure~\ref{introduction} (B), the model continues to learn from outdated supervision that mismatches the current data distribution. Consequently, obsolete decision boundaries are applied to an already shifted feature space, inducing severe negative transfer, a phenomenon termed the blind adaptation zone. In essence, such a supervision gap poses a fundamental challenge to online learning adaptation mechanisms.

In this paper, we propose the \underline{\textbf{D}}ual-\underline{\textbf{T}}rack \underline{\textbf{G}}eometric \underline{\textbf{O}}nline \underline{\textbf{L}}earning (DT-GOL) framework. To enable precise geometric reasoning over streaming data, we first use an online Gaussian copula to project heterogeneous, incomplete data into a unified latent metric space. Subsequently, we construct a dynamic geometric graph to capture the topological evolution of the streaming feature and refine these insights into soft pseudo-labels that perceive uncertainty through a dynamic evidence calibration mechanism. Furthermore, we employ a decoupled dual-track architecture where a master learner anchors stability on verified delayed labels, while a transient branch leverages soft geometric knowledge for low-risk prospective adaptation, effectively bridging the supervision gap.

The main contributions of this paper are as follows:
\begin{itemize}

\item We investigate the challenging problem of nonstationary environment with label delay, a phenomenon that is prevalent in real-world scenarios yet has received scant attention in prior  online learning research.

\item We propose an online learning with novel designs, in which we introduce geometric surrogate learning for prospective adaptation within the delay window and a decoupled dual‑track architecture to balance stability and plasticity.

\item Extensive experiments on real-world and synthetic datasets demonstrate that DT-GOL significantly outperforms state-of-the-art baselines in
nonstationary environment with label delay. Furthermore, we validate that the proposed method exhibits strong adaptability within the blind adaptation zone.
\end{itemize}

\section{RELATED WORK}

Our work is related to the following three research directions, namely online learning with concept drift, delayed feedback learning, and semi-supervised learning.

\subsection{Online Learning with Concept Drift}

The surge in high-speed data streams has made online learning a key paradigm, enabling models to be updated sequentially with each new instance, thus avoiding the high costs of batch retraining \cite{zhou2022open,hoi2021online}. Traditional online learning is set in a stationary environment where the data generation distribution remains constant, and classic algorithms such as OGD~\cite{zinkevich2003online} and FTRL~\cite{mcmahan2013ad} can theoretically achieve sublinear regret bounds and converge to stable optimal solutions, even in cases of feature heterogeneity and missing features~\cite{he2021online}. However, real-world data streams often exhibit non-stationary characteristics, with their underlying data distribution changing over time, this phenomenon is known as concept drift \cite{gaudreault2024systematic}. Our research focuses on the latter, aiming to better reflect the actual complexity of real-time data.

To mitigate concept drift, existing strategies primarily bifurcate into active and passive paradigms. Active methods trigger model adjustments via explicit detection, for example, MCD-DD~\cite{wan2024online} uses contrastive learning to quantify maximum concept discrepancy for boundary identification, while STUDD~\cite{cerqueira2023studd} monitors prediction divergence within a student-teacher architecture. In contrast, passive strategies maintain structural plasticity for continuous adaptation. For instance, CONDOR~\cite{zhao2020handling} dynamically reweights historical model ensembles based on current performance, and OL-MDISF~\cite{zhuo2025online} employs adaptive sliding windows to adjust the learning scope by comparing statistical differences across data sub-windows. However, they are typically limited by the strict assumption of immediate feedback. In scenarios with high validation latency, active detectors fail to detect drift, while passive optimizers lose guidance.

\subsection{Delayed Feedback Learning}
 Delayed feedback learning has garnered significant attention as a critical solution for real-world streaming scenarios where immediate supervision is unattainable~\cite{gomes2022survey}. Existing research primarily follows the approach of addressing validation latency in streaming environments: unbiased statistical estimation and historical information reuse. Statistical methods ($e.g.$, DFM~\cite{pmlr-v28-joulani13}, ~IWMS\cite{csaba2024label}) attempt to rectify loss bias via delay distribution modeling, yet they falter in non-stationary environments where rigid stationarity assumptions disintegrate under volatile dynamics. Conversely, historical reuse frameworks (e.g., HDR~\cite{chan2023capturing}, OTL~\cite{lin2021online}) transfer knowledge from recurring historical patterns but struggle with novel drifts, where enforcing incompatible prototypes induces detrimental misalignment, actively hindering adaptation to emerging distributions. However, these methods often fail when faced with wild streaming data because the calculation of importance weights or similarity metrics becomes unstable.

\subsection{Semi-supervised Learning}
Since the delay window effectively constitutes a stream of unlabeled data, semi-supervised learning offers a promising theoretical foundation naturally by leveraging the manifold assumption to infer supervision from geometric structures~\cite{gomes2022survey,dyer2013compose}. Recently, this paradigm has extended to the field of online learning~\cite{10.1145/3494832}. For example, OLIFL~\cite{10476707} employs an information matrix for dynamic feature weighting and utilizes confidence estimation based on error bounds to filter out false labels, and OSLMF~\cite{wu2023online} utilizes density-peak clustering to propagate supervision from scarce labeled instances to their neighbors in a latent space. However, these methods are designed for label absence rather than label delay, the former referring to permanent disappearance, the latter to delayed arrival. In this paper, we employ semi-supervised learning to continuously model the concept evolution within delayed windows. To provide reliable supervision, we construct geometry-aware soft-pseudo labels. Additionally, we design a decoupled dual-track learning architecture to avoid conflicts between ground-truth and pseudo labels.

\section{PROBLEM STATEMENT}

We investigate the problem of online learning with delayed labels in a non-stationary environment. Formally, let $\mathcal{X} = (\mathbb{R} \cup \{\bot\})^d$ and $\mathcal{Y} = \{0, 1\}$ denote the feature and label spaces, respectively, where $\bot$ represents a missing value. The data stream arrives sequentially as $\mathcal{S}=\{(\mathbf{x}_t, y_t)\}_{t=1}^T$, where each instance $\mathbf{x}_t \in \mathcal{X}$ is drawn from a time-varying distribution $\mathcal{D}_t$ and its ground-truth label $y_t \in \mathcal{Y}$ is revealed after $L$ steps. Thus, at step $t$, the observable history is restricted to $\mathcal{H}_t = \{(\mathbf{x}_i, y_i)\}_{i=1}^{t-L}$ and an unsupervised delay window $\{\mathbf{x}_k\}_{k=t-L+1}^t$. We strictly adhere to the sequential instance-based prediction protocol rather than batch processing.

With the above definitions, the task is defined as follows. At time $t$, the learner receives delayed label $y_{t-L}$, updates parameters from $\theta_{t-1}$ to $\theta_{t}$ based on the verifiable pair $(\mathbf{x}_{t-L}, y_{t-L})$ immediately, and subsequently predicts $\hat{y}_t = f_{\theta_t}(\mathbf{x}_t)$ for the current instance. Despite this update, a critical distributional mismatch persists since the decision boundary $\theta_t$ remains anchored to the historical distribution $D_{t-L}$ while being forced to extrapolate to the potentially drifted current distribution $D_t$. Therefore, we propose to bridge the supervision gap by actively exploiting the topological evolution within the unsupervised window $\{\mathbf{x}_k\}_{k=t-L}^t$ to infer real-time shifts.

\section{Method}

We propose the \underline{D}ual-\underline{T}rack \underline{G}eometric \underline{O}nline \underline{L}earning framework, named DT-GOL. We first introduce an online latent representation learning module that maps wild streaming data into a unified Gaussian space, and then construct a soft pseudo-label generation module to geometrically propagate labels within the delay window while modeling uncertainty. Finally, we implement a delayed online ensemble module that integrates stable ground-truth learning with adaptive pseudo-label learning via a decoupled dual-track architecture.

\subsection{Online Latent Representation Learning}

\subsubsection{Unified Latent Gaussian Mapping}\label{4.1.1}

To realize this unified representation, we employ a Gaussian copula~\cite{zhao2020missing} that maps each observed feature, regardless of its type or missing status, to a common latent variable following a standard Gaussian distribution. Similar latent-space normalizations are often paired with latent factor modeling for high-dimensional and incomplete data, including adaptive divergence, fast autoencoder, fuzzy-PID, particle-swarm, Hessian-vector, and Kalman-filter variants~\cite{10159989,10265117,10502217,10380219,9785520,9839318}.

Formally, let $\mathbf{x}_t = [x_{t,1}, \dots, x_{t,d}]^\top$ be the observed mixed-type input vector at time $t$. We postulate $\mathbf{x}_t \sim GC(\mathbf{\Sigma}, f^{-1})$, which implies the existence of a correlation matrix $\mathbf{\Sigma}$ and a set of element-wise monotone mapping functions $f^{-1} = \{f_1^{-1}, \dots, f_d^{-1}\}$ such that the transformed latent vector $\mathbf{z}_t = f^{-1}(\mathbf{x}_t)$ follows a standard multivariate normal distribution $\mathbf{z}_t \sim \mathcal{N}(\mathbf{0}, \mathbf{\Sigma})$. Each mapping $f_j^{-1}$ is determined by the data type of the $j$-th feature.

For continuous features, $f_j^{-1}$ is strictly invertible, defined as $f_j^{-1} = \Phi^{-1} \circ F_j$, where $F_j$ is the marginal CDF. Conversely, for discrete or ordinal variables where $x_{t,j}\in \{1, \dots, K\}$, the mapping is determined by a set of thresholds $S_j = \{s_{j,1}, \dots, s_{j,K-1}\}$. Let $p_{j,l} = P(x_{t,j}=l)$ be the probability mass of the $l$-th level. The thresholds are defined by the cumulative probabilities:
\begin{equation}
    s_{j,l} = \Phi^{-1}\left(\sum_{r=1}^l p_{j,r}\right).
\end{equation}
 Specifically, for an observed value $k$, the latent variable falls within the interval $(s_{j, k-1}, s_{j, k}]$, with $s_{j,0} = -\infty$ and $s_{j,K} = +\infty$.

In online streaming settings, since the actual marginal distributions $F_j$ and probability masses $p_{j,l}$ are unknown and time-varying, we approximate them dynamically using a coarse buffer $\mathcal{B}_c$, a sliding window, which allows us to adapt to potential distribution shifts in the data stream via the most recent observed instances. For a continuous input $x_{t,j}$, we calculate the empirical CDF with a smoothing factor to ensure finite mapping:\begin{equation}
\hat{F}_j(x_{t,j}) = \frac{1}{|\mathcal{B}_c|+1} \sum_{\mathbf{x}_i \in \mathcal{B}_c} \mathbb{I}(x_{i,j} \leq x_{t,j}),
\end{equation}
where $\mathbb{I}(\cdot)$ denotes the indicator function. The estimated latent value is then obtained via $\hat{z}_{t,j} = \hat{f}_j^{-1}(x_{t,j}) = \Phi^{-1}(\hat{F}_j(x_{t,j}))$.

Similarly, for ordinal features, we estimate the empirical thresholds $\hat{S}_j$ using the frequency counts in $\mathcal{B}_c$, which simplifies to $\mathcal{B}$ in the appendix. The latent representation is then approximated by the expected value of the truncated normal distribution bounded by the empirical intervals $(\hat{s}_{j, k-1}, \hat{s}_{j, k}]$. 

\subsubsection{Online Correlation Learning and Imputation}\label{5.1.3}

After mapping the heterogeneous data to obtain $\mathbf{z}_t = [z_{t,1}, \dots, z_{t,d}]^\top$, our next objective is to capture the dependencies between features via the correlation structure $\mathbf{\Sigma}$, and impute missing entries via an online expectation-maximization (OEM).

The OEM framework described above can adapt to streaming data dynamically, as at each step it incrementally updates the associated structure, and produces a geometrically consistent and interpolated stable representation $\mathbf{z}_t^{\text{rec}}$ (simplified as $\mathbf{z}_t$). This representation is then used to generate soft labels and for ensemble learning, as detailed in Sections~\ref{4.2} and~\ref{4.3}, respectively. Recent work on temporal QoS estimation, tensor compression, sampling-neighborhood regularization, neural nonnegative factorization, PID-controlled learning, battery-life prediction, and traffic imputation further indicates that latent correlation modeling is a flexible substrate for sparse and incomplete streams~\cite{yuan2026novel,li2026adaptive,he2026survey,xu2025sampling,lyu2025genetic,li2025neural,yuan2025proportional,chen2024latent,yang2024latent}.

\subsection{Online Soft Pseudo-label Generation}\label{4.2}

\subsubsection{Geometric Propagation Structure}
\label{4.2.1}

To extract structured knowledge for reliable propagation, we abstract the continuous latent space established into a lightweight topological graph $G = (\mathcal{V}, \mathcal{E})$. Crucially, this continuous representation allows us to capture the smooth temporal evolution of the data stream, a property often lost in discrete or irregular input spaces. Specifically, the vertex set $\mathcal{V} = \{1, \dots, |\mathcal{B}_f|\}$ corresponds to the indices of instances currently held in the buffer $\mathcal{B}_f$, we characterize each instance $\mathbf{z}_i$ ($i \in \mathcal{V}$) using two geometric metrics: the local density $\rho_i$ and the minimum cluster distance $\delta_i$. 

The $\rho_i$ employs a Gaussian kernel to quantify data concentration:
\begin{equation} 
\rho_i = \sum_{\mathbf{z}_j \in \mathcal{V}, j \neq i} \exp\left(-\frac{d^2(\mathbf{z}_i - \mathbf{z}_j)}{d_c^2}\right), 
\end{equation}
where $d(\cdot, \cdot)$ is the Euclidean distance, and $d_c$ is an adaptive cutoff distance, which is typically set to the 1\%-2\% percentile of pairwise distances in $\mathcal{B}_f$~\cite{wu2018self}. The $\delta_i$ identifies density peaks by measuring the distance to the nearest neighbor with higher density:
\begin{equation}
\delta_i =
\begin{cases}
\min\limits_{j \in \mathcal{V},\rho_j > \rho_i} d(\mathbf{z}_i, \mathbf{z}_j), & \text{if } \exists j \in \mathcal{V} \text{ s.t. } \rho_j > \rho_i, \\
\max\limits_{j \in \mathcal{V}} d(\mathbf{z}_i, \mathbf{z}_j), & \text{otherwise}.
\end{cases}
\end{equation}

Based on the derived $(\rho_i,\delta_i)$, we then construct the connectivity mapping $\mathbf{P}: \mathcal{V} \to \mathcal{V} \cup \{\bot\}$ to capture intrinsic density gradients:
\begin{equation}
\mathbf{P}(i) =
\begin{cases}
\underset{j \in \mathcal{V}, \rho_j > \rho_i}{\arg\min} d(\mathbf{z}_i, \mathbf{z}_j), & \text{if } \exists j \in \mathcal{V} \text{ s.t. } \rho_j > \rho_i, \\
\bot, & \text{otherwise},
\end{cases}
\end{equation}
where $\bot$ denotes a local density peak. This mapping establishes the directed edge set $\mathcal{E} = \{(i, j) \mid j = \mathbf{P}(i) \land j \neq \bot\}$, which forms stable geometric pathways for label propagation. This density-guided view is compatible with recent tensor and latent-factor models for dynamic high-dimensional structures, including neural, attention-based, momentum-accelerated, bias-extended, fine-grained regularized, and ADMM-based variants~\cite{lyu2025dynamic,wang2025convolution,lin2025neural,xu2025attention,lin2025momentum,xu2025adaptively,wu2024fine,zhong2024alternating}.

\begin{algorithm}[t]
\small
\caption{Soft Pseudo-Label Self-Training}
\label{alg:propagation}
\SetKwInOut{Input}{Input}
\SetKwInOut{Output}{Output}

\Input{Graph $G$, labeled set $\mathcal{L}$, classifier $f$}
\Output{Updated classifier $f^*$, final soft pseudo-labels $\mathcal{Y}$}

\BlankLine
\tcp{Stage 1: Forward Density-Ascending Propagation}
Initialize active set $\mathcal{S} \leftarrow \mathcal{L}$\;
\While{True}{
    $\mathcal{N}_{+} \leftarrow \mathcal{F}_{+}(\mathcal{S}) \setminus \mathcal{S}$\;
    \If{$\mathcal{N}_{+} = \emptyset$}{
        \textbf{break}\;
    }
    \ForEach{$j \in \mathcal{N}_{+}$}{
        Generate soft pseudo-label $\tilde{y}_j$ via~\ref{al:fusion}\;
    }
    Update $f$ by minimizing loss on batch $\{(\mathbf{z}_j, \tilde{y}_j) \mid j \in \mathcal{N}_{+}\}$\;
    $\mathcal{S} \leftarrow \mathcal{S} \cup \mathcal{N}_{+}$\;
}
$\mathcal{D}_{asc} \leftarrow \mathcal{S} \setminus \mathcal{L}$\;

\BlankLine
\tcp{Stage 2: Backward Density-Descending Propagation}
Initialize active set $\mathcal{R} \leftarrow \mathcal{S}$\;
\While{True}{
    $\mathcal{N}_{-} \leftarrow \mathcal{F}_{-}(\mathcal{R}) \setminus \mathcal{R}$\;
    \If{$\mathcal{N}_{-} = \emptyset$}{
        \textbf{break}\;
    }
    \ForEach{$j \in \mathcal{N}_{-}$}{
        Generate soft pseudo-label $\tilde{y}_j$ via~\ref{al:fusion}\;
    }
    Update $f$ by minimizing loss on batch $\{(\mathbf{z}_j, \tilde{y}_j) \mid j \in \mathcal{N}_{-}\}$\;
    $\mathcal{R} \leftarrow \mathcal{R} \cup \mathcal{N}_{-}$\;
}
$\mathcal{D}_{desc} \leftarrow \mathcal{R} \setminus \mathcal{S}$\;

\BlankLine
\tcp{Stage 3: Global Refinement}
% 整合数据集
Construct augmented training set $\mathcal{T} \leftarrow \mathcal{L} \cup \mathcal{D}_{asc} \cup \mathcal{D}_{desc}$\;
% 全局更新（对应文中 Eq. 6）
Update classifier $f^*$ by minimizing loss on $\mathcal{T}$\;
% 生成最终标签
Compute final pseudo-labels $\mathcal{Y} = \{f^*(\mathbf{z}_i) \mid i \in \mathcal{D}_{asc} \cup \mathcal{D}_{desc}\}$\;

\Return $f^*, \mathcal{Y}$\;
\end{algorithm}

\subsubsection{Soft Pseudo-Label Self-Training}

Based on $G = (\mathcal{V}, \mathcal{E})$ constructed in Section~\ref{4.2.1}, we develop a soft pseudo-label self-training strategy. We first define the density parents $\mathcal{F}_{+}$ and children $\mathcal{F}_{-}$ for any subset $\mathcal{S} \subseteq \mathcal{V}$ to guide the propagation flow:\begin{align}
\mathcal{F}_{+}(\mathcal{S}) &= \{j \in \mathcal{U} \mid \exists i \in \mathcal{S} \text{ s.t. } \mathbf{P}(i) = j\}, \\
\mathcal{F}_{-}(\mathcal{S}) &= \{j \in \mathcal{U} \mid \mathbf{P}(j) \in \mathcal{S}\},
\end{align}
where $\mathcal{U}$ denotes unlabeled remainder. The overall self-training process is summarized in \textbf{Algorithm \ref{alg:propagation}}. As outlined in the algorithm, the process alternates between a forward density-ascending phase and a backward density-descending phase. To ensure robustness during this propagation, we we implement this step using a multi-view evidence fusion strategy rather than rely on a single predictor. Especifically, for each candidate $j$, we synthesize three complementary perspectives $\mathcal{K} = \{ \mathrm{I}, \mathrm{C}, \mathrm{T} \}$:

\begin{itemize}
    \item Intrinsic View ($\mathbf{v}_{\mathrm{I}}$): The instantaneous inference from the current classifier, $\mathbf{v}_{\mathrm{I}} = f(\mathbf{z}_j)$, capturing the model's current discriminatory power.
    
    \item Collective View ($\mathbf{v}_{\mathrm{C}}$): A global prior retrieved from the soft label matrix $\mathbf{M}$, which functions as a noise-resistant memory of class prototypes. We extract the evidence $\mathbf{v}_{\mathrm{C}} = \mathbf{M}_{\hat{y}_j}$ based on the intrinsic hard prediction $\hat{y}_j = \arg\max \mathbf{v}_{\mathrm{I}}$. To ensure purity, $\mathbf{M}$ is dynamically refined via a moving average strictly using verified instances where the model's prediction aligns with the ground truth, thereby effectively filtering out misclassification noise.
        
    \item Neighbor View ($\mathbf{v}_{\mathrm{T}}$):A local geometric consensus derived by aggregating the soft labels of structural neighbors in the current frontier set $\mathcal{N}$ $i.e.,$ $\mathcal{N}_{+}$ or $\mathcal{N}_{-}$, computed as $\mathbf{v}_{\mathrm{T}} = \frac{1}{|\mathcal{N}|} \underset{k \in \mathcal{N}}{\sum} \tilde{y}_k$.
\end{itemize}

 The fused soft label is computed via entropy-based adaptive weighting:
\begin{equation}
\label{al:fusion}
\tilde{y}_j = \frac{\sum\limits_{k \in \mathcal{K}} \alpha_k \mathbf{v}_k}{\sum\limits_{k \in \mathcal{K}} \alpha_k}.
\end{equation}
where dynamic weights $\alpha_k \propto \exp(-H(\mathbf{v}_k) \cdot T)$ are computed based on the Shannon entropy $H(\cdot)$ of each view, effectively prioritizing sharper, less uncertain evidence sources.

\subsection{Online Delayed Ensemble}\label{4.3}

\subsubsection{Progressive Dual-Track Learning}

To structurally decouple robust historical knowledge from volatile pseudo-supervision, we architect an online hierarchical progressive learner, denoted as $\textbf{G}_t^Z = \{G_{\mathcal{P},t}^Z, G_{\mathcal{T},t}^Z\}$, operating within the Gaussian space $Z$, where a persistent learner $G_{\mathcal{P},t}^Z$ and a transient learner $G_{\mathcal{T},t}^Z$ correspond long-term stable learning and short-term adaptation modes, respectively. We adhere to a strict progressive dependency criterion, that is, $G_{\mathcal{T},t}^Z$ is always initialized and constrained based on the state of $G_{\mathcal{P},t}^Z$, thus ensuring that pseudo-labels do not disrupt the stable learning mode and maintain adaptability. Related latent factor optimization studies have explored distributed second-order learning, Nesterov acceleration, Tucker-decomposition recovery, proximal symmetric models, neural Tucker factorization, parallel adaptive SGD, graph-convolutional factor analysis, particle-swarm adjustment, momentum incorporation, and randomized multilayer structures~\cite{wang2024distributed,li2023generalized,mi2023spatio,zhong2023proximal,tang2025auto,qin2023parallel,bi2023two,luo2020position,zhong2020momentum,yuan2020multilayered}.

Specifically, $G_{\mathcal{P},t}^Z$ acts as a stable knowledge backbone that evolves on verified instance-label pairs $(z_{t-L}, y_{t-L})$. Its parameter $w_{\mathcal{P},t}^Z$ undergoes incremental updates via online gradient descent:\begin{equation} w_{\mathcal{P},t}^Z = w_{\mathcal{P},t-1}^Z - \eta \nabla \mathcal{L}(\sigma(w_{\mathcal{P},t-1}^Z \cdot z_{t-L}), y_{t-L}), \end{equation} where $\sigma$ is sigmoid activation function, $\mathcal{L}$ is the cross-entropy loss, and $\eta$ denotes learning rate. The $G_{\mathcal{T},t}^Z$ branches then from $w_{\mathcal{P},t}^Z$ to updated aggressively to the pseudo-labels stream $\mathcal{Q}_t = \{(z_i, \tilde{y}_i) \mid i \in [t-L+1, t-1]\}$, where $\hat{y}_i \in [0, 1]$ represents the soft pseudo-label distribution derived from topological propagation. To fully leverage this fine-grained topological information, we perform a gradient descent step based on the soft cross-entropy loss $\mathcal{L}_{\text{soft}}$:\begin{equation} w_{\mathcal{T},t}^Z = w_{\mathcal{P},t}^Z - \eta \sum_{(z, \tilde{y}) \in \mathcal{Q}_t} \nabla \mathcal{L}_{\text{soft}}(\sigma(w_{\mathcal{P},t}^Z \cdot z_{}), \tilde{y}).\end{equation}
Here, $\mathcal{L}_{\text{soft}}$ measures the distribution gaps to mitigate error propagation. Lastly, $G_{\mathcal{T},t}^Z$ operates under a memoryless protocol, where discarding occurs immediately after the current prediction round, preventing the accumulation of pseudo-label noise.

\begin{table*}[t]
    \centering
    \caption{Average performance comparison under different delays. The shaded areas show the performance results of our proposed DT-GOL. The best results are marked with * and highlighted in bold.}
    \label{AllPerformancecomparison}
    \resizebox{\linewidth}{!}{%
    \begin{tabular}{lcccccccccccccc}
        \toprule
        % --- 第一行表头 ---
        \multicolumn{1}{c}{\multirow{2}{*}[-0.5ex]{\textbf{Dataset}}} & 
        \multicolumn{2}{c}{FOBOS} & 
        \multicolumn{2}{c}{OVFM} & 
        \multicolumn{2}{c}{OSLMF} & 
        \multicolumn{2}{c}{IWMS} & 
        \multicolumn{2}{c}{MDISF} &
        \multicolumn{2}{c}{LACH} & 
        \multicolumn{2}{c}{\cellcolor{mygray}\textbf{DT-GOL}} \\
        
        % --- 调整分割线范围 ---
        \cmidrule(lr){2-3} \cmidrule(lr){4-5} \cmidrule(lr){6-7} \cmidrule(lr){8-9} \cmidrule(lr){10-11} \cmidrule(lr){12-13} \cmidrule(lr){14-15}
        
        % --- 第二行表头 ---
         & CER & AUC & CER & AUC & CER & AUC & CER & AUC & CER & AUC & CER & AUC & \cellcolor{mygray}CER & \cellcolor{mygray}AUC \\
        \midrule
        
        % --- 数据部分：Real Dataset ---
        \multicolumn{15}{l}{\textit{\textbf{Real Dataset}}} \\
        \addlinespace[3pt]
        \hspace{1em}australian & 0.365 & 0.647 & 0.302 & 0.744 & 0.335 & 0.728 & 0.419 & 0.596 & 0.366 & 0.688 & 0.402 & 0.620 & \cellcolor{mygray}\textbf{0.285*} & \cellcolor{mygray}\textbf{0.758*} \\
        
        \hspace{1em}wdbc & 0.388 & 0.878 & 0.129 & 0.920 & 0.152 & 0.839 & 0.184 & 0.847 & 0.164 & 0.824 & 0.366 & 0.598 & \cellcolor{mygray}\textbf{0.129*} & \cellcolor{mygray}\textbf{0.934*} \\
        
        \hspace{1em}wbc & 0.494 & 0.837 & 0.109 & 0.932 & 0.125 & 0.922 & 0.170 & 0.871 & 0.121 & 0.922 & 0.315 & 0.697 & \cellcolor{mygray}\ \textbf{0.075*} & \cellcolor{mygray}\textbf{0.967*} \\
        
        \hspace{1em}ionosphere & 0.360 & 0.617 & 0.337 & 0.664 & 0.387 & 0.619 & 0.379 & 0.581 & 0.453 & 0.535 & 0.333 & 0.665 & \cellcolor{mygray}\textbf{0.325*} & \cellcolor{mygray}\textbf{0.673*} \\

        \hspace{1em}german      & 0.304 & 0.582 & 0.349 & 0.540 & 0.329 & 0.530 & 0.360 & 0.523 & 0.323 & 0.527 & \textbf{0.299*} & \textbf{0.591*} & \cellcolor{mygray}0.338 & \cellcolor{mygray}0.589 \\

        \hspace{1em}diabetes    & 0.379 & 0.422 & 0.367 & 0.591 & 0.359 & 0.609 & 0.361 & 0.524 & 0.357 & 0.610 & 0.347 & 0.620& \cellcolor{mygray}\textbf{0.343*} & \cellcolor{mygray}\textbf{0.620*} \\

        \hspace{1em}credit      & 0.351 & 0.721 & 0.287 & 0.754 & 0.357 & 0.696 & 0.397 & 0.628 & 0.359 & 0.690 & 0.402 & 0.618 & \cellcolor{mygray}\textbf{0.252*} & \cellcolor{mygray}\textbf{0.805*} \\
        
        \hspace{1em}kr-vs-kp      & 0.381 & 0.710 & 0.316 & 0.749 & 0.392 & 0.626 &0.409& 0.605 & 0.399 & 0.618 & 0.418 & 0.618 & \cellcolor{mygray}\textbf{0.311*} & \cellcolor{mygray}\textbf{0.758*} \\
        
        \hspace{1em}chesswaka      & 0.407 & 0.577 & 0.387 & \textbf{0.613*} & 0.419 & 0.525 & 0.479 & 0.549 & 0.397 & 0.575 & \textbf{0.384*} & 0.598 & \cellcolor{mygray}0.384 & \cellcolor{mygray}0.596 \\

        \hspace{1em}LUdata      & 0.405 & 0.663 & 0.312 & 0.725 & 0.382 & 0.669 & 0.421 & 0.608 & 0.384 & 0.663 & 0.401 & 0.630 & \cellcolor{mygray}\textbf{0.307*} & \cellcolor{mygray}\textbf{0.728*}\\

        % --- 数据部分：Synthetic Datasets ---
        \multicolumn{15}{l}{\textit{\textbf{Synthetic Datasets}}} \\
        \addlinespace[3pt]
        \hspace{1em}Agrawal     & 0.464 & 0.535 & 0.427 & 0.593 & 0.434 & 0.571 & 0.444 & 0.554 & 0.455 & 0.547 & 0.411 & \textbf{0.623*} & \cellcolor{mygray}\textbf{0.397*} & \cellcolor{mygray}0.615\\

        \hspace{1em}SEA     & 0.338 & 0.691 & 0.329 & 0.710 & 0.404 & 0.514 & 0.473 & 0.513 & 0.403 & 0.503 & 0.386 & 0.614 & \cellcolor{mygray}\textbf{0.316} & \cellcolor{mygray}\textbf{0.712} \\
        \bottomrule
    \end{tabular}
    }
\end{table*}

\subsubsection{Hierarchical Risk-Aware Integration}\label{4.3.2}

To synthesize a robust global prediction, we employ a hierarchical strategy that stabilizes intra-space predictions prior to performing dynamic inter-space arbitration. We first stabilize intra-space predictions $\hat{p}_t^\mathcal{U}$ for the original ($O$) and latent ($Z$) spaces:
\begin{equation}
    \hat{p}_t^\mathcal{U} = \beta \cdot \sigma(w_{\mathcal{P},t}^\mathcal{U} \cdot z_t^\mathcal{U}) + (1 - \beta) \cdot \sigma(w_{\mathcal{T},t}^\mathcal{U} \cdot z_t^\mathcal{U}),
\end{equation}where $\beta \in [0, 1]$ denotes a fixed reliability coefficient to anchor the transient prediction to the persistent backbone. 

Globally, we aggregate the historical loss on delayed ground truths as $R_t^\mathcal{U} = \sum_{i=1}^{t-L-1} \mathcal{L}(\sigma(w_{\mathcal{P},i}^\mathcal{U} \cdot z_{i}^\mathcal{U}), y_{i})$ to capture the long-term fidelity of each space. The final global prediction $\hat{y}_t$ is computed via a dynamic ensemble, according to the Boltzmann distribution of their historical risks:
\begin{equation}
    \hat{y}_t = \alpha_t \hat{p}_t^Z + (1 - \alpha_t) \hat{p}_t^O, \quad \text{with } \alpha_t = \frac{\exp(-\mu R_t^Z)}{\exp(-\mu R_t^Z) + \exp(-\mu R_t^O)},
\end{equation}
where $\mu=\sqrt{\frac{2\ln 2}{T}}$ and $T$ represents the total time.

\section{EXPERIMENT}

\subsection{Experimental Settings}

\subsubsection{Dataset and Evaluation Metrics.} We evaluate DT-GOL on 14 datasets, comprising 10 real-world datasets and 4 synthetic datasets. Consistent with prior research~\cite{wu2023online,he2021online}, we employ the widely recognized accuracy (ACC) and area under curve (AUC) to assess the overall model performance.

\subsubsection{Baselines and Implementation Details.} We compare our proposed algorithm with five baselines covering standard and delayed settings. Among these baselines, FOBOS~\cite{singer2009efficient}, OVFM~\cite{he2021online}, OSLMF~\cite{wu2023online}, and MDISF~\cite{zhuo2025online} are standard online learning methods that excel at handling complex data streams, but they all assume that labels arrive immediately. In contrast, IWMS~\cite{csaba2024label} and LACH~\cite{10.1145/3662186} represent advanced solutions to the delay label problem, respectively representing data-centric and model-centric approaches. 

\subsection{Overall Performance Comparison}

Table~\ref{AllPerformancecomparison} compares the final CER and AUC performance across all competing methods. Among the baselines, IWMS performs surprisingly poorly, ranking even lower than the classical FOBOS. This suggests that blind adaptation during label delays induces negative transfer and proves less effective than simply waiting for the delayed ground truth. Similarly, MDISF underperforms OSLMF, likely because the label delay disrupts its concept drift adaptation mechanism, further exacerbating negative transfer. The suboptimal performance of OSLMF can be attributed to its inability to distinguish between pseudo-labels and true labels, leading to conflicting gradients that destabilize the optimization landscape and cause oscillations during learning. While LACH shows promise on specific datasets (e.g., German and RBF), it suffers from high variance and poor generalization, as evidenced by significant accuracy drops on datasets such as wdbc. This indicates that while specialized delayed-feedback strategies show promise, ensuring consistent generalization remains a challenge. OVFM emerges as the most robust baseline, securing the second-best average rank in both CER and AUC. Its success stems from probabilistic feature modeling that effectively perceives concept drift and maintains stability, underscoring the potential to address drift by observing changes in feature distributions.

\subsection{Ablation Study}
\begin{table}[t] % [t] 表示优先置顶，符合论文排版习惯
    \centering
    \caption{Ablation studies evaluated by ACC. The best results are marked with * and highlighted in bold.}
    \label{tab:performance_comparison}
    % 使用 resizebox 确保表格严格适应单栏宽度
    \resizebox{\linewidth}{!}{
        \begin{tabular}{lccccc} % l=左对齐(模型名), c=居中对齐(数值)
            \bottomrule
            \bottomrule
            \textbf{Dataset} & DT-GOL & $w/oGC$ & $w/o PL$ & $w/o DT$ & $w/o RA$ \\
            \midrule
            australian  & \textbf{0.715*} & 0.683 & 0.675 & 0.714 & 0.690 \\
            wdbc   & \textbf{0.871*} & 0.869 & 0.845 & 0.843 & 0.844 \\
            wbc   & \textbf{0.925*} & 0.903 & 0.892 & 0.881 & 0.910 \\
            ionosphere   & \textbf{0.675*} & 0.643 & 0.639 & 0.619 & 0.637 \\
            german   & \textbf{0.662*} & 0.639 & 0.632 & 0.627 & 0.635 \\
            diabetes  &  \textbf{0.657*} & 0.639 & 0.638 & 0.642 & 0.667 \\
            credit   & \textbf{0.748*} & 0.744 & 0.717 & 0.725 & 0.716 \\
            kr-vs-kp   & \textbf{0.689*} & 0.687 & 0.687 & 0.575 & 0.612 \\
            chesswaka   & \textbf{0.616*} & 0.595 & 0.602 & 0.601 & 0.602 \\
            LUdata  & \textbf{0.693*} & 0.691 & 0.691 & 0.641 & 0.613 \\
            \midrule
            \textbf{Avg. Drop} & - & 2.26\% $\downarrow$& 3.19\% $\downarrow$& 5.35\% $\downarrow$& 4.57\% $\downarrow$\\
            \bottomrule
            \bottomrule
        \end{tabular}
    }
\end{table}

We evaluate the individual contributions of each module using real-world datasets by constructing four DT-GOL variants for ablation analysis. The first variant, $w/o GC$, removes the Gaussian copula and replaces missing values with zeros. The second, $w/o PL$, eliminates pseudo-labels to ensure the model relies exclusively on delayed-arriving labels within the latent space. The third variant, $w/o RA$, discards the risk-aware ensemble and performs predictions solely through the latent space. Finally, $w/o DT$ removes the dual-track learning mechanism and relies on a single base learner.

Overall, as shown in Table~\ref{tab:performance_comparison}, all components contribute to the model, and their removal leads to performance degradation.  The magnitude of this degradation is consistent across AUC and ACC, $i.e.,$ $w/o DT$ > $w/o RA$ > $w/o PL$ > $w/o GC$, although the magnitudes are relatively small. Specifically, removing the dual-track mechanism ($w/o DT$) causes the most significant performance drop, decreasing by $5.35\%$ and $6.60\%$, respectively. This validates that the decoupled dual-track architecture is critical for preventing error propagation and mitigating gradient conflicts caused by mixed supervisory signals. The risk-aware ensemble strategy ($w/o RA$) also proves essential, yielding average drops of $4.57\%$ and $4.77\%$, indicating that dynamic weighting effectively mitigates prediction variance. Although the performance drops for $w/o PL$ and $w/o GC$ are moderate ($2.26\%$–$3.19\%$), these components remain foundational to the framework's validity, as $w/o GC$ establishes the consistent metric space required for geometric reasoning and $w/o PL$ provides the essential supervision proxy during the label delay window. In summary, DT-GOL consistently outperforms all ablated variants, validating the holistic design of the proposed framework.

\subsection{Performance comparison w.r.t. blind adaptation zone}
\begin{figure}[t]
    \centering
    \begin{minipage}{0.48\linewidth}
        \centering
        \includegraphics[width=\textwidth]{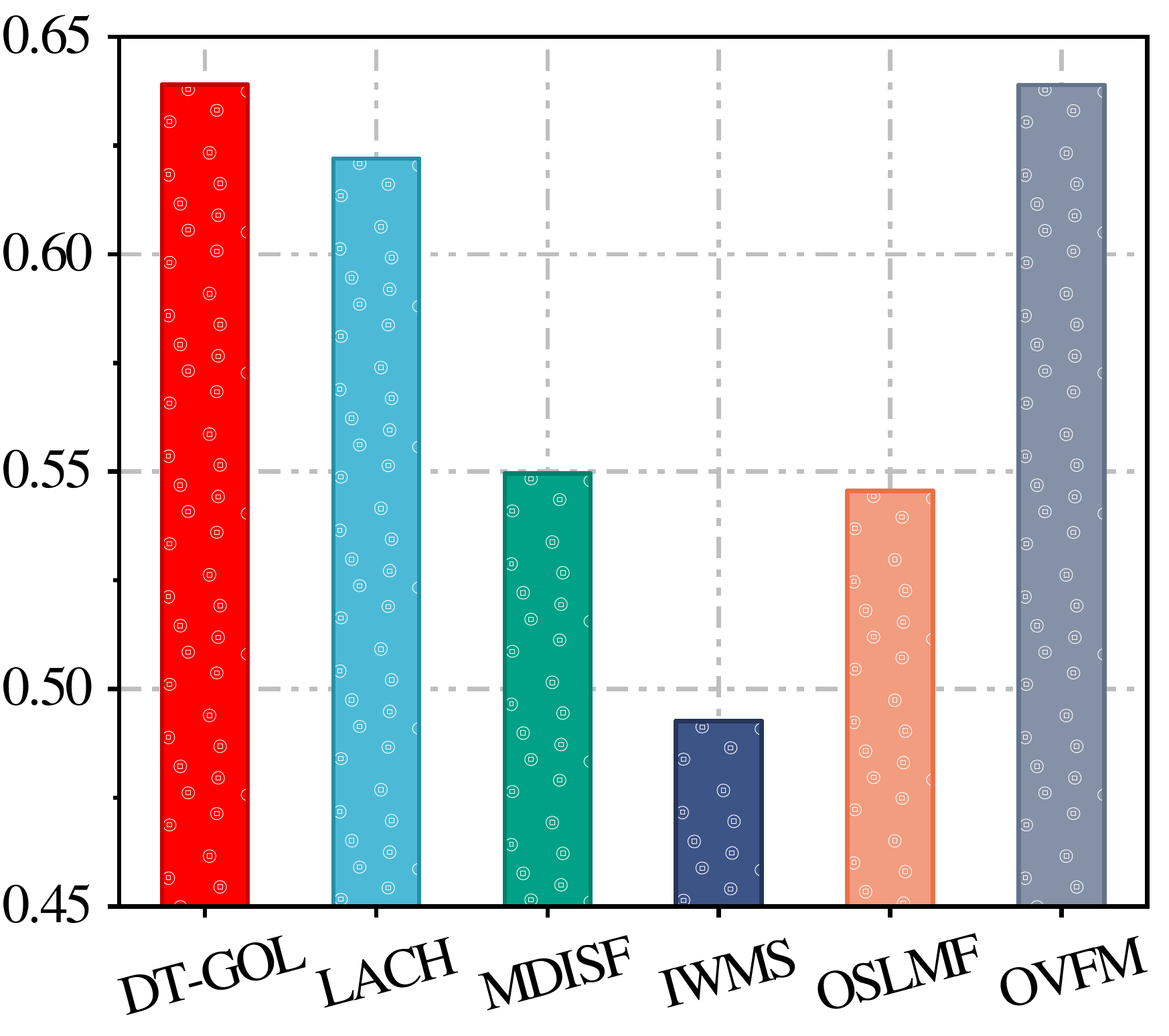}
         \vspace{-0.5em} % 调整此值控制间距
        \label{fig:example_a}
    \end{minipage}
    \hfill
    \begin{minipage}{0.48\linewidth}
        \centering
        \includegraphics[width=\textwidth]{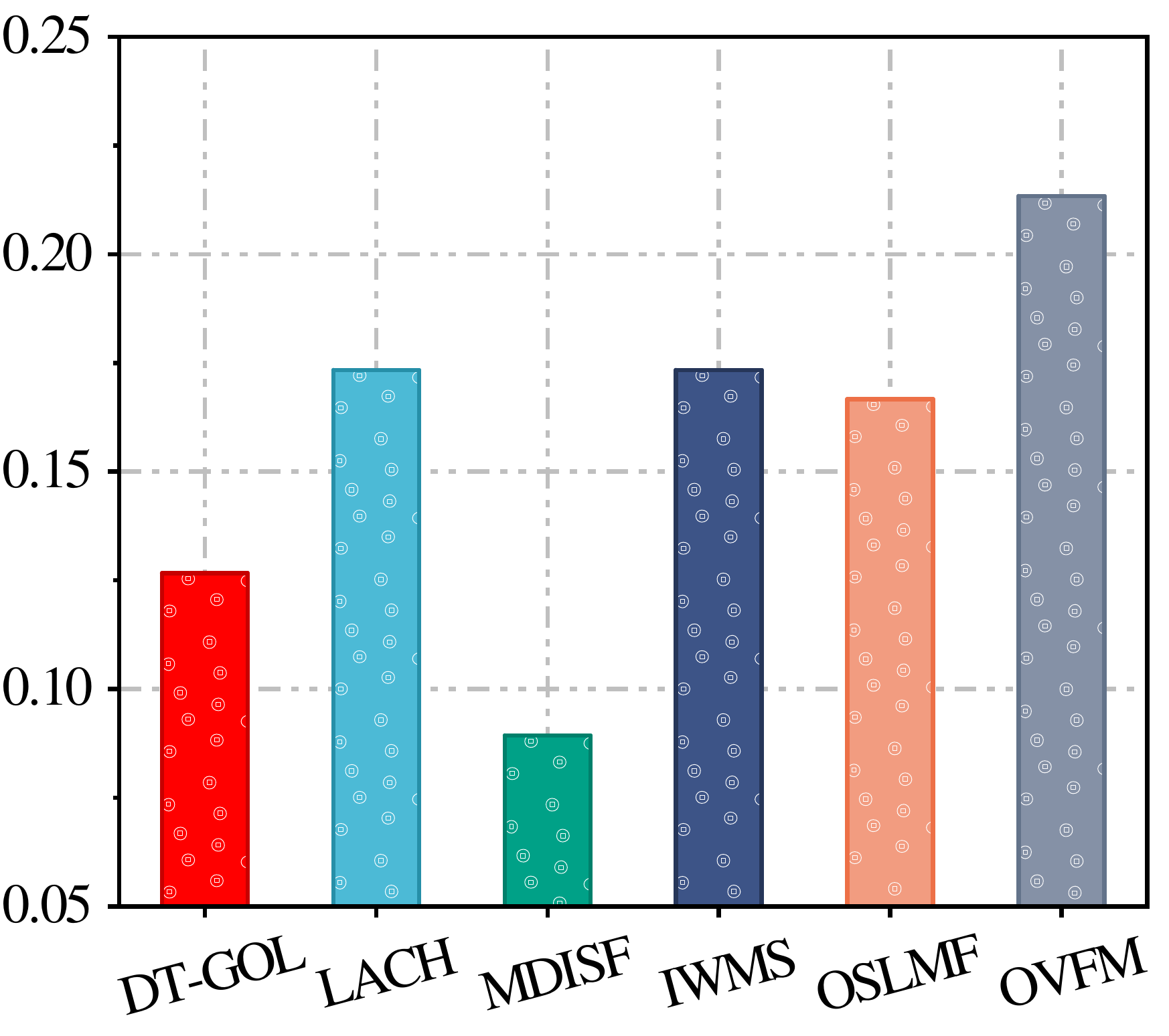}
         \vspace{-0.5em} % 调整此值控制间距
        \label{fig:example_b}
    \end{minipage}
     \vspace{-1em} % 调整此值控制间距
    \caption{Performance analysis in the blind adaptation zone on the SEA dataset.}
    \label{fig:two_figures}
\end{figure}

We evaluate model performance within the blind adaptation zone, defined as the interval $[T, T+\text{delay}]$ between the concept shift at time $T$ and the arrival of updated labels. A synthetic dataset is employed to ensure precise ground truth for shift occurrences. We report average Accuracy (ACC) to measure adaptation capability and maximum performance drop relative to pre-shift levels to quantify robustness.

As shown in Figure~\ref{fig:two_figures}, our method consistently exhibits strong adaptation capability during the blind adaptation zone. In terms of average ACC, our model achieves the highest performance (0.639), slightly surpassing OVFM (0.639) and clearly outperforming other baselines, with improvements ranging from $2.7\%$ to $29.8\%$. Crucially, the right panel highlights the robustness of our approach. While OVFM yields high accuracy, it suffers a severe performance drop (0.213) during the transition. Conversely, MDISF exhibits the lowest drop (0.089) but fails to maintain competitive accuracy (0.550). Our method achieves the optimal trade-off, limiting the maximum drop to 0.127 while maintaining state-of-the-art accuracy. This indicates that our approach effectively balances maintained accuracy and stability in the absence of immediate label feedback.

\bibliographystyle{ACM-Reference-Format}

\end{document}